\def\paperID{637}
\def\confName{ICCV}
\def\confYear{2025}

\def\authorBlock{
    Hao Zhang\textsuperscript{1}\thanks{Equal contribution}
    \quad
    Haolan Xu\textsuperscript{1}\footnotemark[1]
    \quad
    Chun Feng\textsuperscript{1}
    \quad
    Varun Jampani\textsuperscript{2}
    \quad
    Narendra Ahuja\textsuperscript{1} \\
    \textsuperscript{1}University of Illinois Urbana Champaign
    \quad
    \textsuperscript{2}Stability AI\\
    {\tt\small haoz19@illinois.edu}\qquad
    {\tt\small jamesdemon923@gmail.com}\qquad
    {\tt\small chunf2@illinois.edu}\\
    {\tt\small varunjampani@gmail.com}\qquad
    {\tt\small n-ahuja@illinois.edu}\qquad
}

\newif\ifreview 
\newif\ifarxiv 
\newif\ifcamera \newcommand{\cameraready}{\cameratrue}
\newif\ifrebuttal 

\cameraready

\pdfoutput=1
\documentclass[10pt,twocolumn,letterpaper]{article}
\usepackage{bm}
\usepackage{caption}

\ifreview \usepackage[review]{cvpr} \fi
\ifarxiv \usepackage[pagenumbers]{cvpr} \fi
\ifrebuttal \usepackage[rebuttal]{cvpr} \fi
\ifcamera \usepackage{cvpr} \fi

\usepackage{graphicx}	
\usepackage{amsmath}	
\usepackage{amssymb}	
\usepackage{booktabs}
\usepackage{times}
\usepackage{microtype}
\usepackage{epsfig}
\usepackage{caption}
\usepackage{float}
\usepackage{placeins}
\usepackage{color, colortbl}
\usepackage{stfloats}
\usepackage{enumitem}
\usepackage{tabularx}
\usepackage{xstring}
\usepackage{multirow}
\usepackage{xspace}
\usepackage{url}
\usepackage{subcaption}
\usepackage{xcolor}
\usepackage[hang,flushmargin]{footmisc}

\ifcamera \usepackage[accsupp]{axessibility} \fi

% ...

\ifarxiv  \fi

\newcommand{\R}[1]{{%
    \textbf{%
        \ifstrequal{#1}{1}{\textcolor{red}{R#1}}{%
        \ifstrequal{#1}{2}{\textcolor{blue}{R#1}}{%
        \ifstrequal{#1}{3}{\textcolor{magenta}{R#1}}{%
        \ifstrequal{#1}{4}{\textcolor{teal}{R#1}}{%
                           \textcolor{cyan}{R#1}%
        }}}}%
    }%
}}

\usepackage{xr-hyper}

\makeatletter
\newcommand*{\addFileDependency}[1]{
  \typeout{(#1)}
  \@addtofilelist{#1}
  \IfFileExists{#1}{}{\typeout{No file #1.}}
}

\makeatother
\newcommand*{\myexternaldocument}[1]{
    \externaldocument{#1}
    \addFileDependency{#1.tex}
    \addFileDependency{#1.aux}
}

\definecolor{cvprblue}{rgb}{0.21,0.49,0.74}
\usepackage[pagebackref,breaklinks,colorlinks,allcolors=cvprblue]{hyperref}
\usepackage[capitalize]{cleveref}
\crefname{section}{Sec.}{Secs.}
\crefname{table}{Table}{Tables}
\crefname{figure}{Fig.}{Figs.}

\ifarxiv \crefname{appendix}{App.}{Apps.}
\else \crefname{appendix}{Suppl.}{Suppls.} \fi

\unless\ifarxiv \myexternaldocument{_supplementary} \fi

\begin{document}
\title{PhysRig: Differentiable Physics-Based Skinning and Rigging Framework for Realistic Articulated Object Modeling}
\author{\authorBlock}

\maketitle

\begin{abstract}
Skinning and rigging are fundamental components in animation, articulated object reconstruction, motion transfer, and 4D generation. Existing approaches predominantly rely on Linear Blend Skinning (LBS), due to its simplicity and differentiability. However, LBS introduces artifacts such as volume loss and unnatural deformations, and it fails to model elastic materials like soft tissues, fur, and flexible appendages (e.g., elephant trunks, ears, and fatty tissues). In this work, we propose \textbf{PhysRig}: a differentiable physics-based skinning and rigging framework that overcomes these limitations by embedding the rigid skeleton into a volumetric representation (e.g., a tetrahedral mesh), which is simulated as a deformable soft-body structure driven by the animated skeleton. Our method leverages continuum mechanics and discretizes the object as particles embedded in an Eulerian background grid to ensure differentiability with respect to both material properties and skeletal motion. Additionally, we introduce material prototypes, significantly reducing the learning space while maintaining high expressiveness. To evaluate our framework, we construct a comprehensive synthetic dataset using meshes from \textit{Objaverse~\cite{objaverse}, The Amazing Animals Zoo~\cite{zoo}, and MixaMo~\cite{mixamo}}, covering diverse object categories and motion patterns. Our method consistently outperforms traditional LBS-based approaches, generating more realistic and physically plausible results. Furthermore, we demonstrate the applicability of our framework in the pose transfer task highlighting its versatility for articulated object modeling. This project is available at \url{https://physrig.github.io/}.
\end{abstract}
\section{Introduction}
\label{sec:intro}


Skinning and rigging are essential for animating articulated objects and play a critical role in numerous applications, including \textit{character animation, motion retargeting, 4D reconstruction, and generative modeling}. Among existing approaches, \textit{Linear Blend Skinning (LBS)} remains the dominant method due to its efficiency and differentiability. However, LBS suffers from severe limitations, including unnatural distortions (e.g., \textit{collapsing joints, candy-wrapper artifacts, and volume shrinkage}) and an inability to capture the behavior of elastic materials. These artifacts become especially problematic when modeling characters with highly deformable regions, such as an elephant's trunk, a human's soft tissue, or flexible appendages.

To address these shortcomings, we introduce a differentiable physics-based skinning and rigging framework that models articulated object deformation as a volumetric simulation problem. Instead of directly mapping vertices to rigid skeleton transformations, we embed the skeleton into a deformable soft-body volume (e.g., bounded by a set of Gaussians and tetrahedral meshes), which is driven by skeletal motion while respecting fundamental physical principles. In particular, we leverage continuum mechanics and the material point method to establish a fully differentiable deformation process, ensuring that both the material properties and skeletal motion are incorporated in a physically consistent manner. Unlike LBS, which applies simple linear blending, our approach captures intricate material behaviors by modeling stress-strain relationships and dynamic responses to skeletal forces, allowing us to achieve more realistic and physics-driven deformations.

A major challenge encountered with these physics-based methods is a large number of material parameters and complex particle interactions, which makes optimization challenging. To overcome this, we introduce material prototypes, a vocabulary of primitives that can be combined to represent all material properties, and span common deformation behaviors of articulated objects. This novel approach significantly reduces the learning space while maintaining expressiveness.
It provides a structured way to interpolate material properties across different object types, enabling more efficient learning while preserving the diversity of real-world material responses.

Evaluating physics-based skinning models is challenging due to the lack of suitable benchmark datasets. Existing datasets are primarily built via LBS-based deformations and lack sufficient variation in material properties and deformation types. To address this gap, we construct a comprehensive synthetic dataset incorporating meshes from \textit{Objaverse~\cite{objaverse}, The Amazing Animals Zoo~\cite{zoo}, and MixaMo~\cite{mixamo}}, covering a diverse range of objects, motion patterns, and material properties. Using this dataset, we demonstrate that our method outperforms LBS-based approaches, producing more realistic deformations across a variety of articulated objects. Additionally, we showcase the effectiveness of our framework in downstream tasks such as pose transfer and 4D object generation, illustrating its broad applicability.
Our key contributions can be summarized as follows:
\begin{itemize}
    \item \textbf{A differentiable physics-based skinning/rigging framework}, leveraging continuum mechanics to enable realistic and physically plausible deformations while remaining differentiable.
    \item \textbf{A novel material prototype formulation}, which reduces the learning complexity by introducing a structured interpolation approach while maintaining high material expressiveness.
    \item \textbf{A novel synthetic dataset} for evaluating physics-based skinning models, demonstrating our framework’s superiority over LBS-based approaches.
\end{itemize}

Our approach bridges the gap between physics-based simulation and differentiable learning, providing a powerful tool for articulated object modeling in computer vision and graphics. By introducing a differentiable physics-driven deformation process, our framework enables new opportunities for more accurate, physically consistent skinning and rigging, with broad implications for animation, motion generation, and 4D modeling.

\section{Related Work}
\label{sec:related}

\textbf{Skinning in 4D Modeling and Animation.}
Skinning is fundamental to 3D character animation, modeling surface deformations induced by skeletal motion~\cite{Cmfaya, Beacsfe}. Among various techniques, Linear Blend Skinning (LBS) remains the most widely used due to its simplicity and computational efficiency~\cite{PSD}. 

LBS is integral to many vision tasks, including video-to-3D reconstruction and avatar modeling. Parametric models like SCAPE~\cite{SCAPE} and SMPL~\cite{smpl} rely on predefined skeletons and skinning weights, limiting their adaptability. Neural implicit approaches~\cite{lasr, viser, NPMs, banmo, nphm, gart, learning, s3o, opennerf} improve generalization but still require precise skeletal information. In avatar modeling, explicit methods~\cite{econ, selfrecon, ghum, point} optimize SMPL parameters, whereas implicit ones~\cite{humansplat, 3dgs, hugs, gauhuman, humannerf, hnerf, anerf} leverage neural representations but face challenges in optimization and topological consistency. 

LBS has also been applied to pose transfer~\cite{3dpt, sfpt}, with MagicPose4D~\cite{mp4d} enabling cross-species motion. However, these methods often require recalculating skeletons and skinning weights for novel motions. Since LBS linearly blends external skeletal motion, it fails to capture true internal deformations, prompting research into physically-based skinning~\cite{soft, fast, effi, pbcs}. While such methods better model volumetric changes, their non-differentiability limits integration with deep learning. Our approach introduces a differentiable physics-based skinning model, enabling efficient joint optimization via gradient descent.

\textbf{Physical 4D Generation.}
In multiphysics simulation, the Material Point Method (MPM)~\cite{moving, affine, material} excels in handling topology changes and frictional interactions across various materials. Recent works~\cite{physgaussian, simany, phys4dgen, fu2024sync4d} integrate MPM for physically plausible motion but rely on manual parameter tuning. Differentiable approaches~\cite{physdreamer, dreamphysics, physics3d} learn material properties but are restricted to simple motions. To bridge this gap, we propose PhysRig, a differentiable physics framework that learns material parameters for articulated objects, ensuring physical consistency across complex motions.

\begin{figure*}[ht]
    \centering
    \includegraphics[width=1\textwidth]{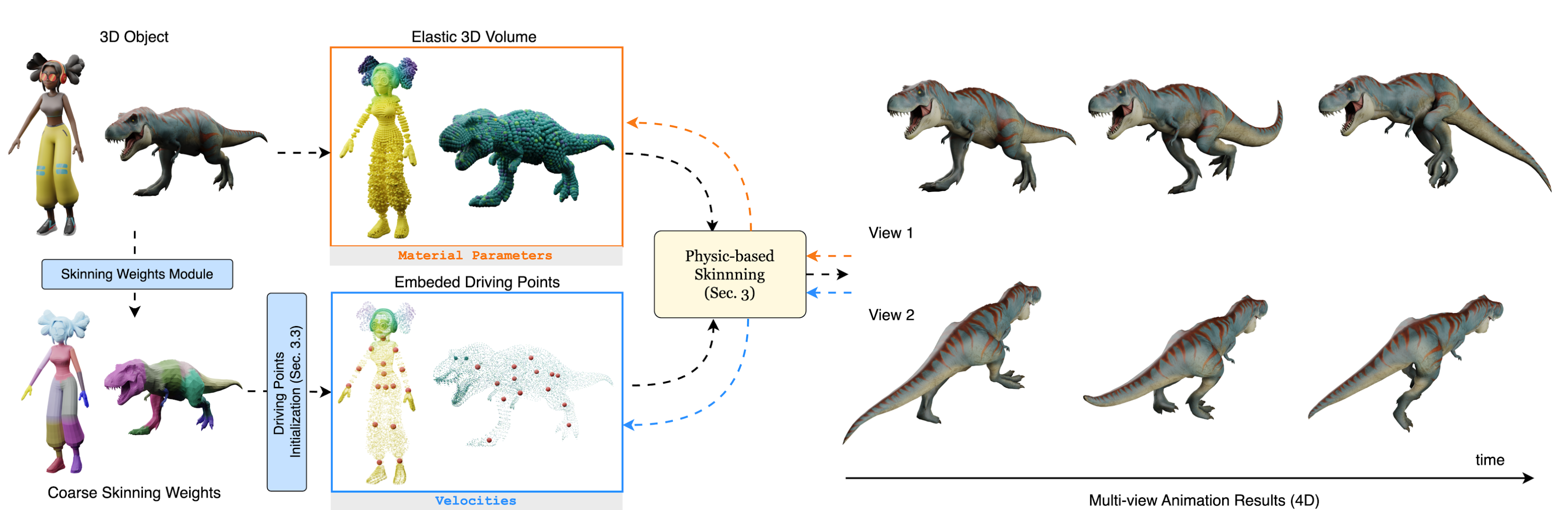}
    \caption{Overview of \textbf{PhysRig}. Given a 3D object, we first compute coarse skinning weights, which initialize embedded driving points for local deformation control. These points, assigned velocities, are linked to an elastic 3D volume with material parameters governing deformation. The differentiable physics-based skinning module generates natural deformations, optimizing velocities and material properties via backward propagation. Finally, multi-view animations illustrate physically plausible shape deformations over time.}
    \vspace{-0pt}
    \label{overview}
\end{figure*}

\section{Method}
\label{sec:method} 
In this paper, we introduce PhysRig, a differentiable physics-based skinning framework for 3D object deformation, applicable to meshes, point clouds, and Gaussian representations. If the input is a mesh or a Gaussian representation, we first perform a filling operation to obtain a solid volume in the form of a point cloud, and the total number of points is \( N \).
As shown in Fig.~\ref{overview}, unlike traditional Linear Blend Skinning (LBS), which applies a weighted sum of bone transformations, PhysRig employs a differentiable physics simulator (Sec.\ref{physics}) to model the 3D object as a volumetric structure. Instead of directly manipulating vertex positions, it embeds driving points within the volume to induce deformation.
PhysRig optimizes two key components to achieve fine-grained control and produce the desired deformation:
\begin{itemize}
    \item \textbf{Material properties}, including Young’s modulus \( E \in \mathbb{R}^P \) and Poisson’s ratio \( \mathbf{\nu} \in \mathbb{R}^P \) for \( P \) material prototypes. The material properties of all \( N \) points are then computed using a function based on the Mahalanobis distance between each point and the material prototypes (Sec.~\ref{mat_prototype}). These properties govern elasticity and deformation behavior, determining how internal forces propagate through the structure.

    \item \textbf{Driving point velocities}, \( v \in \mathbb{R}^{\{l*M, 3\}} \), representing the motion of the internal skeletal structure parameterized by transformations \( \{\mathbf{t}_0, ..., \mathbf{t}_M\}, \mathbf{t}_i \in SE(3) \), where \( M \) is the number of virtual joints. These velocities $v$ drive the deformation, with \( l = 8 \) set by default and the driving points' positions are initialized by coarse skinning weights or uniform sampling (detailed in Sec.~\ref{drive_point}).
\end{itemize}
The driving points encode the skeletal motion, propagating movement to the surrounding 3D volume, while the material properties define how internal motion influences the object's outer surface. PhysRig can be formulated as:
\begin{equation}
    \mathbf{X}' = \mathcal{F}(\mathbf{X}, E, \nu, v, \Delta t),
\end{equation}
where \( \mathbf{X} \in \mathbb{R}^{N \times 3} \) denotes the initial point positions, and \( \Delta t \in \mathbb{R} \) is the time step governing temporal evolution. The function \( \mathcal{F} \) computes the deformed positions \( \mathbf{X}' \) via a differentiable physics simulation.

\subsection{Physics-Based Simulation}
\label{physics}

To model object deformations under external interactions, we simulate motion using the principles of continuum mechanics. Our approach represents objects as continuous volumetric materials governed by conservation laws, enabling differentiable physics-based deformation modeling.

\subsubsection{Continuum Mechanics Formulation}

We describe the motion of a deformable object using a time-dependent mapping function \( \phi \), which transforms material coordinates \( \bm{X} \) in the undeformed space \( \Omega_0 \) to world coordinates \( \bm{x} \) in the deformed space \( \Omega_t \):
$\bm{x} = \phi(\bm{X}, t).$
The evolution of \( \phi \) is constrained by fundamental physical laws:

\textbf{Conservation of Mass.} The total mass within a material region remains constant over time:

\begin{equation}
    \int_{B^{t}_{\epsilon}} \rho(\bm{x}, t)\,d\bm{x} = \int_{B^{0}_{\epsilon}} \rho(\phi^{-1}(\bm{x}, t), 0)\,d\bm{x},
\end{equation}

where \( \rho(\bm{x}, t) \) is the density field.

\textbf{Conservation of Momentum.} The motion of the object is dictated by the balance of internal and external forces:

\begin{equation}
    \int_{B^{t}_{\epsilon}} \rho(\bm{x}, t) \bm{a}(\bm{x}, t) \,d\bm{x} = \int_{\partial B^{t}_{\epsilon}} \sigma \cdot \bm{n} \, d\bm{x} + \int_{B^{t}_{\epsilon}} \bm{f}^{\text{ext}} \,d\bm{x},
\end{equation}

where \( \bm{a}(\bm{x}, t) = \frac{\partial^{2} \phi}{\partial t^{2}} \) represents acceleration, \( \bm{f}^{\text{ext}} \) denotes external forces, and \( \sigma \) is the Cauchy stress tensor, which encodes local deformation behavior.

\subsubsection{Material Model and Deformation Representation}

To model elastic responses, we use a constitutive model relating the stress tensor \( \sigma \) to the deformation gradient \( F = \frac{\partial \phi}{\partial \bm{X}} \). We adopt a Fixed Corotated hyperelastic material model, which effectively captures nonlinear deformations while maintaining stability.

The Cauchy stress tensor is derived from the strain energy density function \( \psi(F) \):

\begin{equation}
    \sigma = \frac{1}{\det(F)} \frac{\partial \psi}{\partial F} F^{T}.
\end{equation}

Following the Fixed Corotated model, the strain energy function is given by:

\begin{equation}
    \psi(F) = \mu \sum_{i=1}^{d} (\sigma_i - 1)^2 + \frac{\lambda}{2} (\det(F) - 1)^2,
\end{equation}

where \( \sigma_i \) are the singular values of \( F \), and the material parameters \( \mu \) and \( \lambda \) are related to Young’s modulus \( E \) and Poisson’s ratio \( \nu \):

\begin{equation}
    \mu = \frac{E}{2(1+\nu)}, \quad \lambda = \frac{E \nu}{(1+\nu)(1-2\nu)}.
\end{equation}

\subsubsection{Simulation via the Material Point Method}

We employ the Material Point Method (MPM)~\cite{moving} to solve the governing equations efficiently. MPM discretizes the object as particles embedded in an Eulerian background grid, enabling robust handling of large deformations while ensuring differentiability.

\textbf{Particle-to-Grid (P2G) Transfer.}  
At each simulation step, per-particle mass and momentum are transferred to the grid using B-spline interpolation:
\begin{equation}
\begin{aligned}
    m_i v_i &= \sum_p N(x_i - x_p) \bigg[ m_p v_p + \Big( m_p C_p \\
    &\quad - \frac{4}{\Delta x^2 \Delta t} V_p \frac{\partial \psi}{\partial F} F_p^T \Big) (x_i - x_p) \bigg] + f_i.
\end{aligned}
\end{equation}
where:
\( m_i \), \( v_i \) are the mass and velocity at grid node \( i \),
\( N(x_i - x_p) \) is the interpolation kernel,
\( C_p \) is the velocity gradient at the particle,
\( f_i \) is the external force,
\( V_p \) is the volume of the particle, which scales the contribution of the stress force term.
The stress-based force term,
$V_p \frac{\partial \psi}{\partial F} F_p^T,$
represents the internal elastic forces exerted by the particle. The factor \( V_p \) ensures that the contribution is properly scaled according to the physical size of the particle, preventing instabilities when transferring forces to the grid.

\textbf{Grid-to-Particle (G2P) Update.}  
After computing velocity updates on the grid, the velocities are interpolated back to particles, and positions are updated:

\begin{equation}
    v_p^{t+1} = \sum_i N(x_i - x_p) v_i, \quad x_p^{t+1} = x_p + \Delta t v_p^{t+1}.
\end{equation}

\textbf{Deformation Gradient Update.}  
The velocity gradient and deformation gradient are updated as:
\begin{equation}
\begin{aligned}
    \nabla v_p^{t+1} &= \frac{4}{\Delta x^2} \sum_i N(x_i - x_p) v_i (x_i - x_p)^T, \\
    F_p^{t+1} &= (I + \Delta t \sum_i v_i \nabla N(x_i - x_p)^T) F_p.
\end{aligned}
\end{equation}
where:
\( \nabla v_p^{t+1} \) is the velocity gradient at particle \( p \), describing how velocity varies locally.
\( F_p^{t+1} \) is the updated deformation gradient, tracking material deformation over time.
\( \nabla N(x_i - x_p) \) is the spatial gradient of the interpolation function, describing how interpolation weights change with position.
\( I \) is the identity matrix, ensuring that the deformation gradient starts from an undeformed state.
\( \Delta t \) is the time step, controlling how much deformation accumulates per iteration.
By iterating these updates, MPM efficiently captures complex material deformations while maintaining differentiability.

\textbf{Driving Points Gradient Update.}
Driving points influence the motion of specific object regions by modifying the velocities of nearby grid nodes within their control region. The velocity update for a driving point \( v_{\text{d}, j} \) is determined by the contributions from the affected grid nodes and is given by:
\begin{equation}
    v_{\text{d}, j} \leftarrow v_{\text{d}, j} + \frac{1}{|R_c|} \sum_{i\in R_c} \nabla v_{i},
\end{equation}
where \( \nabla v_{i} \) represents the velocity gradient at grid node \( i \) within the control region $R_c$.

\begin{table*}[h]
    \centering
    \resizebox{\textwidth}{!}{%
    \begin{tabular}{c|cc|cc|cc|cc|cc|cc|cc|cc|cc}
        \toprule
        \multirow{3}{*}{Method} 
        & \multicolumn{10}{c|}{Humanoid Character} 
        & \multicolumn{8}{c}{Quadruped Animal} \\
        & \multicolumn{2}{c}{Michelle} & \multicolumn{2}{c}{Ortiz} 
        & \multicolumn{2}{c}{Mutant} & \multicolumn{2}{c}{Jellyman} 
        & \multicolumn{2}{c|}{Kaya}  
        & \multicolumn{2}{c}{Leopard} & \multicolumn{2}{c}{Mammoth} 
        & \multicolumn{2}{c}{Stego} & \multicolumn{2}{c}{Krin} \\
        & UR $\uparrow$ & CD $\downarrow$ & UR $\uparrow$ & CD $\downarrow$ 
        & UR $\uparrow$ & CD $\downarrow$ & UR $\uparrow$ & CD $\downarrow$ 
        & UR $\uparrow$ & CD $\downarrow$  
        & UR $\uparrow$ & CD $\downarrow$ & UR $\uparrow$ & CD $\downarrow$ 
        & UR $\uparrow$ & CD $\downarrow$ & UR $\uparrow$ & CD $\downarrow$ \\
        \midrule
        LBS-1 & 2.82 & 1.426 & 2.18 & 2.993 & 2.02 & 2.512 & 3.05 & 6.532 & 2.42 & 2.081 & 3.03 & 2.413 & 2.54 & 1.782 & 3.07 & 0.682 & 3.03 & 0.561 \\
        LBS-2 & 3.06 & 0.891 & 2.93 & 2.011 & 3.37 & 1.438 & 3.25 & 4.130 & 2.97 & 1.214 & 3.13 & 1.328 & 2.8 & 0.891 & 3.64 & 0.202 & 3.38 & 0.271 \\
        LBS-3 & 3.32 & 0.372 & 3.01 & 1.472 & 3.47 & 0.801 & 3.66 & 3.811 & 3.43 & 0.408 & 3.21 & 0.493 & 3.25 & 0.346 & 3.83 & 0.103 & 3.47 & 0.048 \\
        Ours-init & 3.19 & 2.105 & 2.91 & 12.755 & 2.98 & 5.977 & 3.43 & 14.161 & 3.39 & 2.027 & 3.43 & 1.822 & 3.13 & 4.991 & 3.51 & 1.664 & 3.63 & 0.844 \\
        Ours & \textbf{4.7} & \textbf{0.139} & \textbf{4.43} & \textbf{1.375} & \textbf{4.61} & \textbf{0.527} & \textbf{4.34} & \textbf{1.214} & \textbf{4.8} & \textbf{0.228} & \textbf{4.45} & \textbf{0.212} & \textbf{4.59} & \textbf{0.127} & \textbf{4.5} & \textbf{0.085} & \textbf{4.4} & \textbf{0.032} \\
        \midrule
    \end{tabular}
    }
    \vspace{-0.45cm}
    \label{tab:rigging_comparison}
\end{table*}
\begin{table*}[h]
    \centering
    \resizebox{\textwidth}{!}{%
    \begin{tabular}{c|cc|cc|cc|cc|cc|cc|cc|cc|cc}
            \multirow{3}{*}{Method} 
            & \multicolumn{4}{c|}{Quadruped Animal} 
            & \multicolumn{14}{c}{Other Entities} \\
            & \multicolumn{2}{c}{Cow} & \multicolumn{2}{c|}{Raccoon} 
            & \multicolumn{2}{c}{T-rex} & \multicolumn{2}{c}{Pterosaur} 
            & \multicolumn{2}{c}{Whale}  
            & \multicolumn{2}{c}{Angelfish} & \multicolumn{2}{c}{Cobra} 
            & \multicolumn{2}{c}{Shark} & \multicolumn{2}{c}{Ave.} \\
            & UR $\uparrow$ & CD $\downarrow$ & UR $\uparrow$ & CD $\downarrow$ 
            & UR $\uparrow$ & CD $\downarrow$ & UR $\uparrow$ & CD $\downarrow$ 
            & UR $\uparrow$ & CD $\downarrow$  
            & UR $\uparrow$ & CD $\downarrow$ & UR $\uparrow$ & CD $\downarrow$ 
            & UR $\uparrow$ & CD $\downarrow$ & UR $\uparrow$ & CD $\downarrow$ \\
            \midrule
            LBS-1 & 2.51 & 1.351 & 2.29 & 1.243 & 2.88 & 5.861 & 2.54 & 6.722 & 2.77 & 0.292 & 3.11 & 0.841 & 2.64 & 2.712 & 3.27 & 0.078 & 2.72 & 2.43 \\
            LBS-2 & 3.0 & 1.019 & 2.58 & 0.791 & 3.21 & 3.763 & 2.78 & 4.512 & 2.95 & 0.253 & 3.57 & 0.614 & 2.79 & 2.133 & 3.59 & 0.046 & 3.12 & 1.56 \\
            LBS-3 & 2.99 & 0.781 & 3.48 & 0.342 & 3.25 & 0.687 & 3.05 & 1.082 & 3.17 & 0.134 & 3.61 & 0.209 & 3.01 & 0.607 & 3.81 & 0.031 & 3.35 & 0.73 \\
            Ours-init & 3.1 & 2.244 & 3.37 & 4.737 & 3.17 & 6.574 & 2.97 & 9.522 & 3.11 & 4.561 & 3.81 & 0.296 & 3.77 & 4.049 & 3.93 & 0.178 & 3.34 & 4.67 \\
            Ours & \textbf{4.24} & \textbf{0.187} & \textbf{4.63} & \textbf{0.198} & \textbf{4.66} & \textbf{0.588} & \textbf{4.49} & \textbf{0.653} & \textbf{4.56} & \textbf{0.132} & \textbf{4.32} & \textbf{0.021} & \textbf{4.76} & \textbf{0.372} & \textbf{4.34} & \textbf{0.016} & \textbf{4.52} & \textbf{0.37} \\  

            \bottomrule
        \end{tabular}
    }
    \caption{\textbf{Comparison of different rigging methods for inverse skinning.} 
    UR $\uparrow$: User Study Rate, CD $\downarrow$: Chamfer Distance. LBS-1, LBS-2, and LBS-3 correspond to using RigNet~\cite{rignet}, Pinocchio~\cite{pinocchio}, and ground truth skinning weights, respectively, as initialization for jointly optimizing skinning weights and bone transformation. \textbf{PhysRig} utilizes Pinocchio to obtain coarse skinning weights for initializing driving points, and then iteratively learns material parameters and driving point velocities. Our dataset consists of 17 diverse objects among humans, quadrupeds, and other entities, totaling 120 motion sequences. We report the average performance across all motions for objects with multiple motions. The User Study setup is provided in the appendix Sec. A.4}
    \vspace{-0pt}
    \label{inverse_skinning}
\end{table*}

\subsubsection{Optimization Strategy for Inverse Skinning}

\textbf{Inverse Skinning} is the process of recovering underlying motion parameters, such as material properties and driving point velocities, from observed deformations of a 3D object. Unlike traditional skinning methods (LBS), where deformations are computed from transformations and skinning weights, our inverse skinning aims to estimate the driving point velocities \( v \) and material properties (Young’s modulus \( E \), Poisson’s ratio \( \nu \)) that best explain a given motion sequence. This requires optimizing physical parameters to minimize discrepancies between simulated and observed motion.

\textbf{Iteritively Optimization.} To ensure stability, we adopt an iterative training strategy. First, we initialize the positions of the driving points and estimate their approximate velocities for each frame. We then alternate between the following two optimization steps:  
(1) Material Parameter Optimization: Fix the driving point velocities and update the material parameters using all frames as a single batch.  
(2) Driving Point Velocity Optimization: Fix the material parameters and sequentially update the velocities of the driving points for each frame. The optimization progresses frame by frame, moving to the next frame once the loss falls below a predefined threshold.  
These two steps are repeated iteratively until either the overall loss falls below a set threshold or the total number of iterations reaches the stopping criterion.

This strategy is designed to account for the differing requirements of material parameters and velocity optimization. Optimizing material parameters requires information accumulated across multiple frames, as the material properties influence the object's global behavior over time. In contrast, optimizing driving point velocities must be performed sequentially on a per-frame basis. Simultaneously optimizing velocities across multiple frames is ineffective, as accurate simulation of later frames is only meaningful if the preceding frames have already been well-optimized.

\subsection{Material Prototype}
\label{mat_prototype}
To efficiently represent material properties across an object's volume, we introduce material prototypes, each characterized by two learnable parameters: Young’s modulus and Poisson’s ratio. The material properties at any point within the volume are computed as a weighted sum of these prototypes. The weights are determined using a function based on the \textbf{Mahalanobis distance} between the query position and the prototypes. Specifically, we define each \textit{material prototype} as a Gaussian ellipsoid, parameterized by its center $\mathbf{C} \in \mathbb{R}^{P \times 3}$, orientation $\mathbf{V} \in \mathbb{R}^{P \times 3 \times 3}$, and diagonal scale $\boldsymbol{\Lambda} \in \mathbb{R}^{P \times 3 \times 3}$, where $P$ denotes the number of prototypes. The weight assignment follows:
\begin{equation}
    W_{n,p} = \text{softmax}_{p \in P}(d(\mathbf{x}_n, \mathbf{C}_{p}, \mathbf{Q}_p))
\end{equation}
where $d(\mathbf{x}_n, \mathbf{C}_p, \mathbf{Q}_p)$ is the Mahalanobis distance, defined as:
$d(\mathbf{x}_n, \mathbf{C}_p, \mathbf{Q}_p) = (\mathbf{x}_n - \mathbf{C}_p)^T \mathbf{Q}_p (\mathbf{x}_n - \mathbf{C}_p), \quad \mathbf{Q}_p = \mathbf{V}_p^T \boldsymbol{\Lambda}_p \mathbf{V}_p.$
Here, $\mathbf{x}_n$ represents the coordinates of a query point $n$, and the Mahalanobis distance function ensures that weights are assigned based on the spatial relationship between the query position and the material prototypes. This formulation enables an efficient and differentiable material representation that generalizes across diverse volumetric structures.

Compared to directly learning per-point material properties or employing a triplane-based function that maps spatial coordinates to material parameters, our material prototype representation offers a significantly more compact and efficient parameterization. By leveraging a small set of prototypes rather than densely modeling every point, we substantially reduce the optimization space while maintaining high expressiveness. Moreover, the prototype-based formulation naturally enforces smooth material transitions, preventing noisy or abrupt variations that are common in per-point learning approaches. This property aligns more closely with the behavior of real-world materials, where material properties exhibit gradual spatial variations rather than sharp discontinuities. 

\subsection{Driving Point Initialization}
\label{drive_point}

Driving points are a crucial component of PhysRig, as efficiently initializing their positions and velocities significantly improves optimization efficiency. To achieve this, we propose a coarse-to-fine initialization strategy based on skinning weights. 
We first obtain coarse skinning weights using existing rigging models such as Pinocchio~\cite{pinocchio} or RigNet~\cite{rignet}, which provide an approximate mapping between the object's surface and skeletal structure. We then place driving points at joint locations, which naturally reside at the boundaries between adjacent parts.

\subsubsection{Affinity-Based Seg via Spectral Clustering}
Given per-vertex skinning weights $\mathbf{W} \in \mathbb{R}^{N \times B}$, where $N$ is the number of vertices and $B$ is the number of bones, we construct an affinity matrix $\mathbf{A}$ to measure similarity between vertices:
\begin{equation}
    A_{i,j} = \exp\left(-\frac{\|\mathbf{W}_i - \mathbf{W}_j\|^2}{\sigma^2}\right),
\end{equation}
where $\sigma$ controls the sensitivity of similarity measurement. A larger $\sigma$ results in smoother clustering, while a smaller $\sigma$ captures finer-scale differences. Using $\mathbf{A}$, we compute the graph Laplacian:
$\mathbf{L} = \mathbf{D} - \mathbf{A}$,where $D_{i,i} = \sum_j A_{i,j}.$
We obtain a low-dimensional embedding by computing the \textit{k} smallest eigenvectors of $\mathbf{L}$ and apply $k$-means clustering to segment the object into rigid regions, each assigned a cluster label $c_i$. Note that $k$ could be different with $B$. Since the coarse skinning weights may not always meet our expectations, our approach allows for flexible control over the number of parts by adjusting $k$.

\subsubsection{Locating Joint via Skinning Weight Variance}
To extract joint locations, we analyze the segmentation output to identify transition regions where adjacent rigid components meet. A vertex $i$ is classified as a \textbf{boundary vertex} if:
$c_i \neq c_j, \quad \text{for some } j \in \mathcal{N}(i),$
where $\mathcal{N}(i)$ denotes the set of neighboring vertices in the mesh. These boundary vertices form the primary candidates for joint locations.
To further refine the detected joints, we analyze \textbf{variance in skinning weights} at boundary vertices. Specifically, we define the joint set $\mathcal{J}$ as:
$\mathcal{J} = \left\{ i \mid \sum_{b} \left( W_{i,b} - \bar{W}_{\mathcal{N}(i),b} \right)^2 > \tau \right\},$
where $\bar{W}_{\mathcal{N}(i),b}$ is the mean skinning weight of neighboring vertices of $i$, and $\tau$ is a threshold for detecting significant weight variations. This step ensures that only regions with meaningful changes in skinning influences are selected as joints.

\subsubsection{Driving Points Initialization}

At each identified \textbf{joint}, we uniformly place $l$ driving points to ensure fine-grained control over the deformation of nearby volumetric regions. Each driving point’s initial velocity is computed as the \textbf{average velocity of its surrounding volume}, ensuring a smooth and physically consistent initialization:
$v_p = \frac{\sum_{i \in \mathcal{N}_p} v_i}{|\mathcal{N}_p|},$
where $\mathcal{N}_p$ represents the set of nearby points influencing the driving point.

Although the coarse skinning weights obtained from pre-existing models may not be highly accurate, they provide a decent starting point. Our method refines these initial estimates (velocity) during optimization, ultimately yielding more accurate motion parameters that adapt to the specific material properties of the object.

\begin{figure}[ht]
    \centering
    \includegraphics[width=0.43\textwidth]{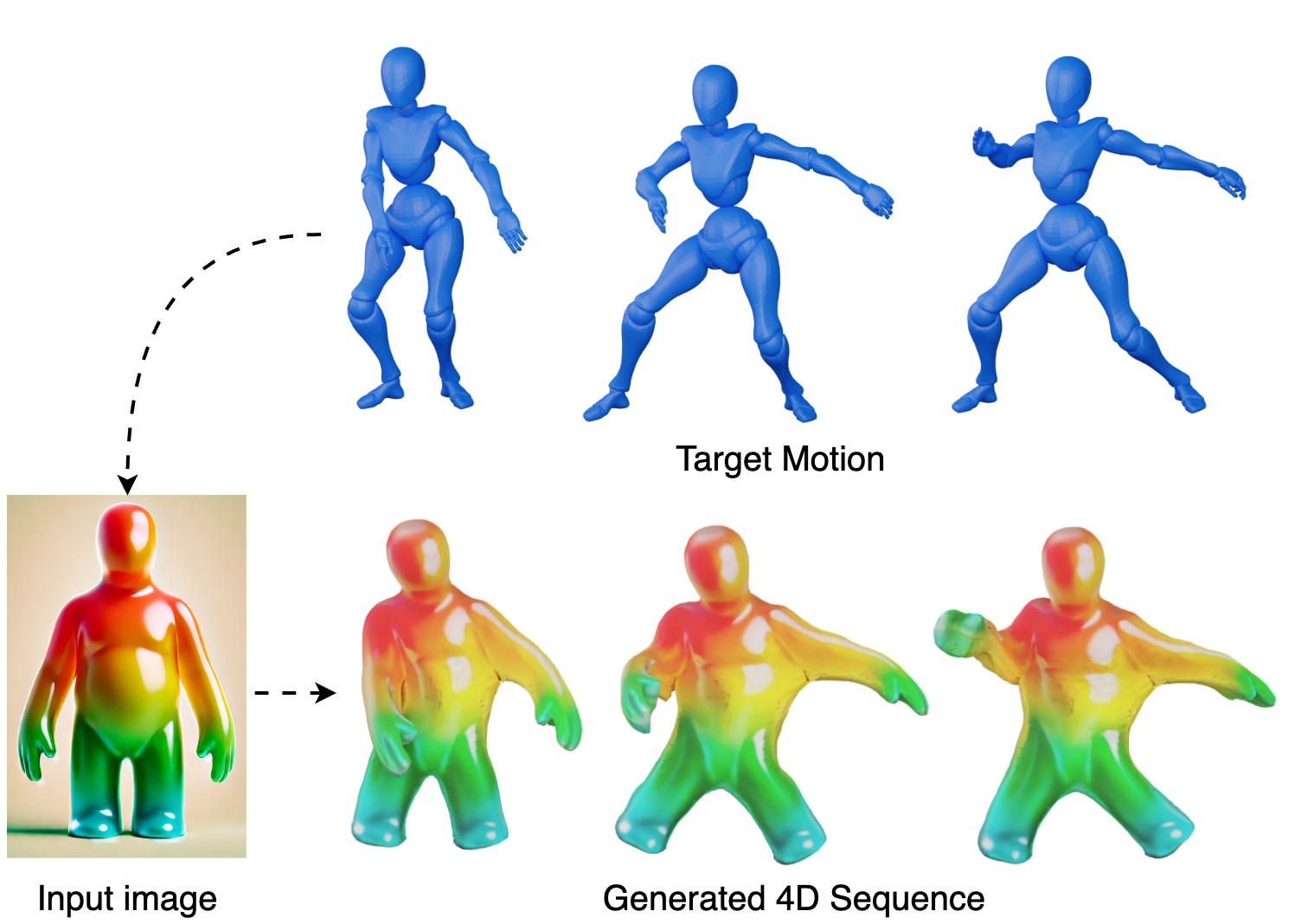}
    \caption{\textbf{PhysRig} enables pose transfer for generated objects.}
    \label{pose_transfer}
\end{figure}

\section{Experiments}

\begin{figure*}[ht]
    \centering
    \includegraphics[width=1\textwidth]{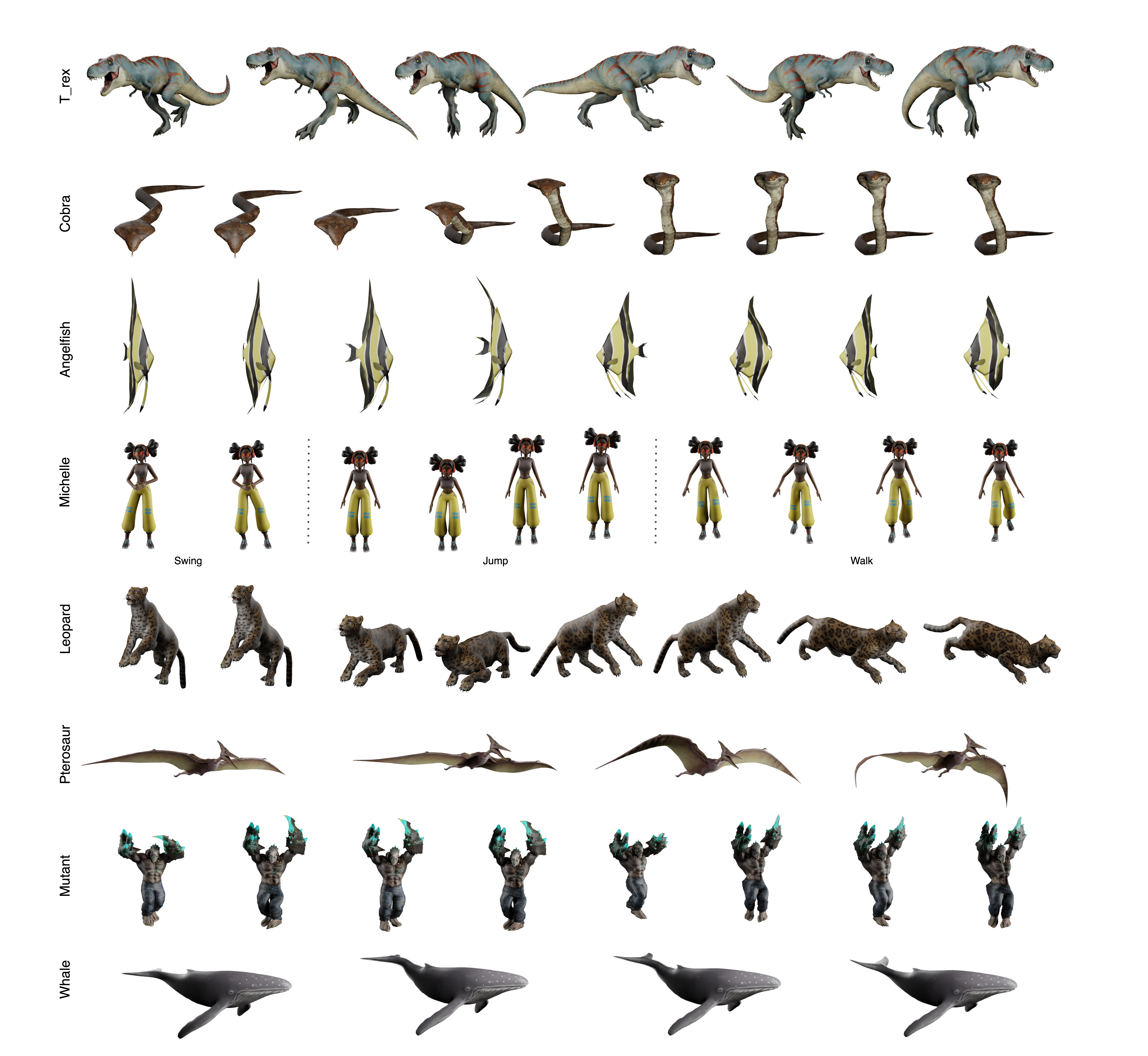}
    \vspace{-5pt}
    \caption{\textbf{Animation results} from the \textbf{PhysRig} approach. These results are obtained from the inverse skinning problem by optimizing material properties and driving point velocities to minimize the deviation from the ground truth mesh sequence.}
    \vspace{-5pt}
    \label{exp_1}
\end{figure*}

\label{experiments}

\begin{table}[]
\centering
\resizebox{0.475\textwidth}{!}{%
\begin{tabular}{c|cc|cc|cc|c}
\hline
\multirow{2}{*}{} &
\multicolumn{2}{c|}{Michelle} &
\multicolumn{2}{c|}{Leopard} &
\multicolumn{2}{c|}{Angelfish} &
\multirow{2}{*}{\begin{tabular}[c]{@{}c@{}}Converge \\ Iteration$\downarrow$\end{tabular}} \\
& UR$\uparrow$ & CD$\downarrow$ & UR$\uparrow$ & CD$\downarrow$ & UR$\uparrow$ & CD$\downarrow$ & \\ \hline
Mat Field & 3.31 & 1.93 & 3.47 & 1.58 & 3.93 & 0.23 & - \\
Per-point & 3.15 & 2.31 & 3.61 & 1.77 & 3.73 & 0.25 & - \\
w/o Locating & 4.31 & 0.186 & 4.23 & 0.358 & 3.97 & 0.031 & 8000 \\
w/o Vel Init & 4.08 & 0.183 & 4.03 & 0.301 & 4.17 & 0.029 & 5000 \\
Prototypes: 25 & - & 0.147 & - & 0.229 & - & 0.023 & \textbf{2000} \\
Prototypes: 100 & \textbf{4.7} & 0.139 & \textbf{4.45} & 0.212 & \textbf{4.32} & 0.021 & 2500 \\
Prototypes: 200 & - & \textbf{0.133} & - & \textbf{0.207} & - & \textbf{0.019} & 2500 \\ \hline
\end{tabular}}
\caption{\textbf{Ablation study} on material prototypes vs. material field vs. per-point for material representation, the impact of the number of material prototypes, and the effect of driving point initialization, including (i) joint localization and (ii) velocity initialization.}
\label{ablation}
\end{table}

In this section, we compare PhysRig with the traditional neural Linear Blend Skinning (LBS) method on the inverse skinning task, which serves as a fundamental component for various applications such as 3D video reconstruction and part decomposition. This comparison highlights PhysRig's strong capability in dynamic modeling and optimization for articulated objects. 
To facilitate the evaluation, we introduce a new dataset, which is constructed from existing datasets (Objaverse, The Amazing Animals Zoo and Mixamo) and includes entities with diverse structural variations. Additionally, we generate a large amount of synthetic data using PhysRig, enabling a more comprehensive analysis of its optimization performance, particularly in learning material properties and driving point velocities.  
For more details on the dataset (Sec. A.1) and implementation (Sec. A.2), more experimental (Sec. A.3) results, and video results please refer to the supplementary materials.

\begin{figure*}[ht]
    \centering
    \includegraphics[width=1\textwidth]{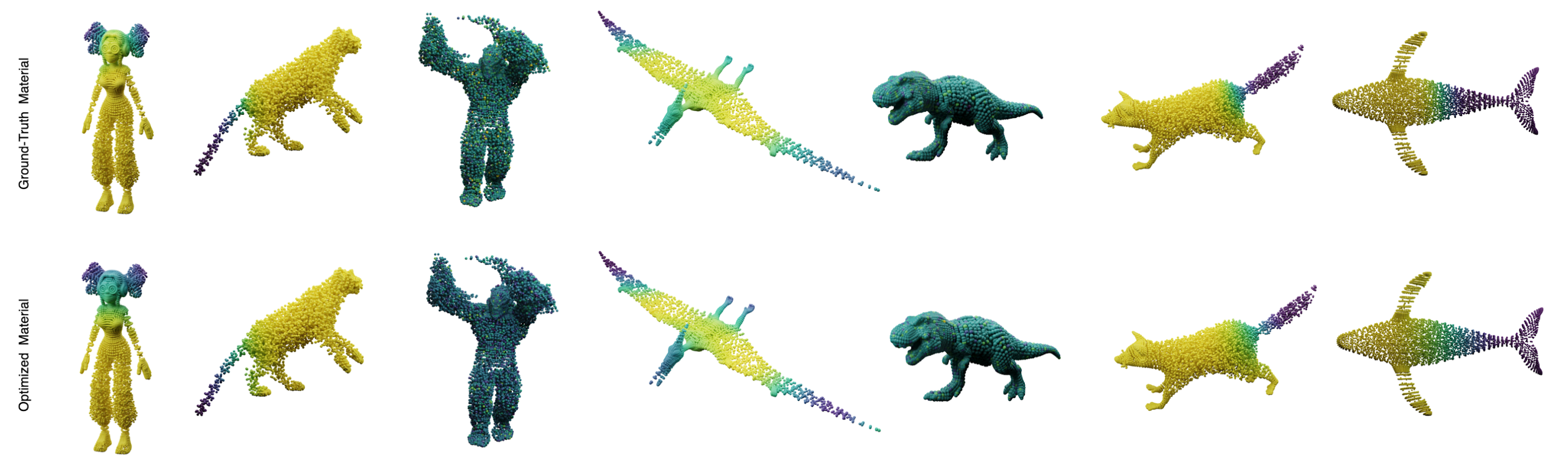}
    \caption{Comparison of the learned material properties with ground truth using our method.}
    \vspace{-5pt}
    \label{exp_2}
\end{figure*}

\subsection{Inverse Skinning Evaluation}

We evaluate the effectiveness of our inverse skinning method across a diverse set of humanoid characters, quadruped animals, and other articulated entities. We compare against traditional Linear Blend Skinning (LBS) baselines, including RigNet~\cite{rignet}, Pinocchio~\cite{pinocchio}, and ground truth skinning weight initialization, as well as the results after driving points initialization before optimization (Ours-init). The evaluation metrics include \textbf{User Study Rate (UR)}, which quantifies perceptual quality based on user preferences, with scores ranging from 0 to 5 (higher is better), and \textbf{Chamfer Distance (CD)}, which evaluates geometric fidelity.

Table~\ref{inverse_skinning} presents the results. Our method consistently outperforms all baselines, achieving the highest UR scores and the lowest CD across all evaluated categories. Notably, on humanoid characters, our method achieves a UR of 4.7 on Michelle and a UR of 4.8 on Kaya, surpassing all LBS-based approaches. Similarly, for quadrupeds, our approach demonstrates superior performance, particularly on the Leopard (UR: 4.45, CD: 0.212) and Stego (UR: 4.5, CD: 0.085), highlighting its robustness across diverse morphologies.
Our approach also generalizes well to other articulated entities, such as the Angelfish (UR: 4.32, CD: 0.021) and Pterosaur (UR: 4.49, CD: 0.653), showcasing its effectiveness beyond conventional character rigging. These results indicate that our inverse skinning formulation not only improves perceptual quality but also significantly reduces geometric error compared to existing baselines. For qualitative comparisons and analysis, please refer to the appendix (Sec. A.3).

\subsection{Ablation Study}

We conduct an ablation study to analyze the impact of different components in our method, particularly focusing on material representation and driving point initialization. The results are summarized in Table~\ref{ablation}.

\textbf{Material Representation:} We compare our prototype-based material representation against material fields and per-point assignments. The per-point approach leads to higher geometric error (CD: 2.31 on Michelle, 1.77 on Leopard), indicating that it struggles to find the optimal solution. The material field (triplane) method also underperforms, demonstrating increased Chamfer Distance across all test cases. In contrast, our prototype-based representation significantly reduces CD and achieves high UR scores.

\textbf{Effect of Driving Point Initialization:} Removing joint localization (w/o Locating) results in increased CD values (e.g., 0.186 on Michelle), requiring 8000 iterations for convergence. Similarly, excluding velocity initialization (w/o Vel Init) leads to a higher CD (0.183 on Michelle) and slower convergence (5000 iterations). These findings suggest that both joint localization and velocity initialization are crucial for improving optimization efficiency and accuracy.

\textbf{Effect of Material Prototype Count:} We also investigate the impact of the number of material prototypes. Reducing the prototype count to 25 does not degrade performance and instead accelerates convergence, achieving the fastest convergence at 2000 iterations while maintaining competitive accuracy (CD: 0.147 on Michelle). Increasing the prototype count to 100 strikes a good balance between performance and convergence time (UR: 4.7, CD: 0.139 on Michelle, convergence: 2500 iterations). Further increasing the prototypes to 200 yields a marginal improvement in CD (0.133 on Michelle) but does not significantly affect UR, suggesting diminishing returns.

Overall, these results demonstrate that our material prototype representation, combined with joint localization and velocity initialization, leads to improved inverse skinning accuracy and faster optimization convergence.

\subsection{Apply PhysRig for Pose Transfering}
\label{posetransfer}
As shown in Figure \ref{pose_transfer}, PhysRig enables pose transfer by taking a mesh sequence as input. Inspired by MagicPose4D~\cite{mp4d}, we first extract the skeleton from the input mesh and align it with the generated mesh. By transferring the bone angles at each frame, we obtain the skeleton sequence for the generated object. This allows us to compute joint velocities between consecutive frames, which serve as the driving point velocities for deforming the generated mesh (volume).
Unlike traditional methods that rely on skinning weight, PhysRig achieves more realistic deformations while significantly improving generalization, as it eliminates the need for explicit skinning weight prediction.

\section{Conclusion}
\label{sec:conclusion}
We introduced PhysRig, a differentiable physics-based skinning framework that addresses the limitations of Linear Blend Skinning (LBS) by modeling deformations through volumetric simulation. By embedding skeletons into a soft-body representation and leveraging continuum mechanics, our approach achieves realistic, physically plausible deformations while remaining fully differentiable.
To enhance efficiency, we introduced material prototypes, reducing learning complexity while maintaining expressiveness. Our evaluation of a diverse synthetic dataset demonstrated superior performance over traditional LBS-based methods. Additionally, PhysRig enables applications such as pose transfer, motion retargeting, and 4D generation, bridging the gap between physics-based simulation and differentiable learning.
Future work includes integrating real-world priors and optimizing for real-time applications, expanding PhysRig's potential in animation and simulation.

\clearpage
{\small
\bibliographystyle{ieeenat_fullname}
\bibliography{11_references}

\begin{thebibliography}{47}
\providecommand{\natexlab}[1]{#1}
\providecommand{\url}[1]{\texttt{#1}}
\expandafter\ifx\csname urlstyle\endcsname\relax
  \providecommand{\doi}[1]{doi: #1}\else
  \providecommand{\doi}{doi: \begingroup \urlstyle{rm}\Url}\fi

\bibitem[{Adobe}()]{mixamo}
{Adobe}.
\newblock {Mixamo}.

\bibitem[Anguelov et~al.(2005)Anguelov, Srinivasan, Koller, Thrun, Rodgers, and Davis]{SCAPE}
Dragomir Anguelov, Praveen Srinivasan, Daphne Koller, Sebastian Thrun, Jim Rodgers, and James Davis.
\newblock Scape: shape completion and animation of people.
\newblock \emph{ACM Trans. Graph.}, 24\penalty0 (3):\penalty0 408–416, 2005.

\bibitem[Baran and Popovi\'{c}(2007)]{pinocchio}
Ilya Baran and Jovan Popovi\'{c}.
\newblock Automatic rigging and animation of 3d characters.
\newblock 26\penalty0 (3):\penalty0 72–es, 2007.

\bibitem[Deitke et~al.(2023)Deitke, Liu, Wallingford, Ngo, Michel, Kusupati, Fan, Laforte, Voleti, Gadre, et~al.]{objaverse}
Matt Deitke, Ruoshi Liu, Matthew Wallingford, Huong Ngo, Oscar Michel, Aditya Kusupati, Alan Fan, Christian Laforte, Vikram Voleti, Samir~Yitzhak Gadre, et~al.
\newblock Objaverse-xl: A universe of 10m+ 3d objects.
\newblock \emph{Advances in Neural Information Processing Systems}, 36:\penalty0 35799--35813, 2023.

\bibitem[Fu et~al.(2024)Fu, Wei, Shen, Song, Yang, Liu, Yang, and Lin]{fu2024sync4d}
Zhoujie Fu, Jiacheng Wei, Wenhao Shen, Chaoyue Song, Xiaofeng Yang, Fayao Liu, Xulei Yang, and Guosheng Lin.
\newblock Sync4d: Video guided controllable dynamics for physics-based 4d generation.
\newblock \emph{arXiv preprint arXiv:2405.16849}, 2024.

\bibitem[Galoppo et~al.(2007)Galoppo, Otaduy, Tekin, Gross, and Lin]{soft}
Nico Galoppo, Miguel~A Otaduy, Serhat Tekin, Markus Gross, and Ming~C Lin.
\newblock Soft articulated characters with fast contact handling.
\newblock In \emph{Computer Graphics Forum}, pages 243--253. Wiley Online Library, 2007.

\bibitem[Giebenhain et~al.(2023)Giebenhain, Kirschstein, Georgopoulos, R{\"{u}}nz, Agapito, and Nie{\ss}ner]{nphm}
Simon Giebenhain, Tobias Kirschstein, Markos Georgopoulos, Martin R{\"{u}}nz, Lourdes Agapito, and Matthias Nie{\ss}ner.
\newblock Learning neural parametric head models.
\newblock In \emph{Proc. IEEE Conf. on Computer Vision and Pattern Recognition (CVPR)}, 2023.

\bibitem[Hu et~al.(2024)Hu, Hu, and Liu]{gauhuman}
Shoukang Hu, Tao Hu, and Ziwei Liu.
\newblock Gauhuman: Articulated gaussian splatting from monocular human videos.
\newblock In \emph{Proceedings of the IEEE/CVF conference on computer vision and pattern recognition}, pages 20418--20431, 2024.

\bibitem[Hu et~al.(2018)Hu, Fang, Ge, Qu, Zhu, Pradhana, and Jiang]{moving}
Yuanming Hu, Yu Fang, Ziheng Ge, Ziyin Qu, Yixin Zhu, Andre Pradhana, and Chenfanfu Jiang.
\newblock A moving least squares material point method with displacement discontinuity and two-way rigid body coupling.
\newblock \emph{ACM Transactions on Graphics (TOG)}, 37\penalty0 (4):\penalty0 1--14, 2018.

\bibitem[Huang et~al.(2024)Huang, Zhang, Zeng, Zhang, Li, Zuo, and Lau]{dreamphysics}
Tianyu Huang, Haoze Zhang, Yihan Zeng, Zhilu Zhang, Hui Li, Wangmeng Zuo, and Rynson~WH Lau.
\newblock Dreamphysics: Learning physical properties of dynamic 3d gaussians with video diffusion priors.
\newblock \emph{arXiv preprint arXiv:2406.01476}, 2024.

\bibitem[Jiang et~al.(2022)Jiang, Hong, Bao, and Zhang]{selfrecon}
Boyi Jiang, Yang Hong, Hujun Bao, and Juyong Zhang.
\newblock Selfrecon: Self reconstruction your digital avatar from monocular video.
\newblock In \emph{Proceedings of the IEEE/CVF Conference on Computer Vision and Pattern Recognition}, pages 5605--5615, 2022.

\bibitem[Jiang et~al.(2015)Jiang, Schroeder, Selle, Teran, and Stomakhin]{affine}
Chenfanfu Jiang, Craig Schroeder, Andrew Selle, Joseph Teran, and Alexey Stomakhin.
\newblock The affine particle-in-cell method.
\newblock \emph{ACM Transactions on Graphics (TOG)}, 34\penalty0 (4):\penalty0 1--10, 2015.

\bibitem[Jiang et~al.(2016)Jiang, Schroeder, Teran, Stomakhin, and Selle]{material}
Chenfanfu Jiang, Craig Schroeder, Joseph Teran, Alexey Stomakhin, and Andrew Selle.
\newblock The material point method for simulating continuum materials.
\newblock In \emph{Acm siggraph 2016 courses}, pages 1--52. 2016.

\bibitem[Kim and Pollard(2011)]{fast}
Junggon Kim and Nancy~S Pollard.
\newblock Fast simulation of skeleton-driven deformable body characters.
\newblock \emph{ACM Transactions on Graphics (TOG)}, 30\penalty0 (5):\penalty0 1--19, 2011.

\bibitem[Kim and James(2011)]{pbcs}
Theodore Kim and Doug~L James.
\newblock Physics-based character skinning using multi-domain subspace deformations.
\newblock In \emph{Proceedings of the 2011 ACM SIGGRAPH/eurographics symposium on computer animation}, pages 63--72, 2011.

\bibitem[Kocabas et~al.(2024)Kocabas, Chang, Gabriel, Tuzel, and Ranjan]{hugs}
Muhammed Kocabas, Jen-Hao~Rick Chang, James Gabriel, Oncel Tuzel, and Anurag Ranjan.
\newblock Hugs: Human gaussian splats.
\newblock In \emph{Proceedings of the IEEE/CVF conference on computer vision and pattern recognition}, pages 505--515, 2024.

\bibitem[Lei et~al.(2024)Lei, Wang, Pavlakos, Liu, and Daniilidis]{gart}
Jiahui Lei, Yufu Wang, Georgios Pavlakos, Lingjie Liu, and Kostas Daniilidis.
\newblock Gart: Gaussian articulated template models.
\newblock In \emph{Proceedings of the IEEE/CVF conference on computer vision and pattern recognition}, pages 19876--19887, 2024.

\bibitem[Lewis et~al.(2000)Lewis, Cordner, and Fong]{PSD}
J.~P. Lewis, Matt Cordner, and Nickson Fong.
\newblock Pose space deformation: a unified approach to shape interpolation and skeleton-driven deformation.
\newblock In \emph{Proceedings of the 27th Annual Conference on Computer Graphics and Interactive Techniques}, page 165–172, USA, 2000. ACM Press/Addison-Wesley Publishing Co.

\bibitem[Liao et~al.(2022)Liao, Yang, Saito, Pons-Moll, and Zhou]{sfpt}
Zhouyingcheng Liao, Jimei Yang, Jun Saito, Gerard Pons-Moll, and Yang Zhou.
\newblock Skeleton-free pose transfer for stylized 3d characters.
\newblock In \emph{European Conference on Computer Vision}, pages 640--656. Springer, 2022.

\bibitem[Lin et~al.(2024)Lin, Wang, Jiang, Hou, and Jiang]{phys4dgen}
Jiajing Lin, Zhenzhong Wang, Shu Jiang, Yongjie Hou, and Min Jiang.
\newblock Phys4dgen: A physics-driven framework for controllable and efficient 4d content generation from a single image.
\newblock \emph{arXiv preprint arXiv:2411.16800}, 2024.

\bibitem[Liu et~al.(2024)Liu, Wang, Yao, Zhang, Zhou, and Duan]{physics3d}
Fangfu Liu, Hanyang Wang, Shunyu Yao, Shengjun Zhang, Jie Zhou, and Yueqi Duan.
\newblock Physics3d: Learning physical properties of 3d gaussians via video diffusion.
\newblock \emph{arXiv preprint arXiv:2406.04338}, 2024.

\bibitem[Loper et~al.(2015)Loper, Mahmood, Romero, Pons-Moll, and Black]{smpl}
Matthew Loper, Naureen Mahmood, Javier Romero, Gerard Pons-Moll, and Michael~J. Black.
\newblock Smpl: a skinned multi-person linear model.
\newblock \emph{ACM Trans. Graph.}, 34\penalty0 (6), 2015.

\bibitem[Magnenat-Thalmann and Thalmann(1991)]{Cmfaya}
N. Magnenat-Thalmann and D. Thalmann.
\newblock Complex models for animating synthetic actors.
\newblock \emph{IEEE Computer Graphics and Applications}, 11\penalty0 (5):\penalty0 32--44, 1991.

\bibitem[McAdams et~al.(2011)McAdams, Zhu, Selle, Empey, Tamstorf, Teran, and Sifakis]{effi}
Aleka McAdams, Yongning Zhu, Andrew Selle, Mark Empey, Rasmus Tamstorf, Joseph Teran, and Eftychios Sifakis.
\newblock Efficient elasticity for character skinning with contact and collisions.
\newblock In \emph{ACM SIGGRAPH 2011 papers}, pages 1--12. 2011.

\bibitem[Mohr and Gleicher(2003)]{Beacsfe}
Alex Mohr and Michael Gleicher.
\newblock Building efficient, accurate character skins from examples.
\newblock \emph{ACM Trans. Graph.}, 22\penalty0 (3):\penalty0 562–568, 2003.

\bibitem[Palafox et~al.(2021)Palafox, Bo{\v z}i{\v c}, Thies, Nie{\ss}ner, and Dai]{NPMs}
Pablo Palafox, Alja{\v z} Bo{\v z}i{\v c}, Justus Thies, Matthias Nie{\ss}ner, and Angela Dai.
\newblock Npms: Neural parametric models for 3d deformable shapes.
\newblock In \emph{Proceedings - 2021 IEEE/CVF International Conference on Computer Vision, ICCV 2021}, pages 12675--12685. Institute of Electrical and Electronics Engineers Inc., 2021.
\newblock Publisher Copyright: {\textcopyright} 2021 IEEE; 18th IEEE/CVF International Conference on Computer Vision, ICCV 2021 ; Conference date: 11-10-2021 Through 17-10-2021.

\bibitem[Pan et~al.(2024)Pan, Su, Lin, Fan, Zhang, Li, Shen, Mu, and Liu]{humansplat}
Panwang Pan, Zhuo Su, Chenguo Lin, Zhen Fan, Yongjie Zhang, Zeming Li, Tingting Shen, Yadong Mu, and Yebin Liu.
\newblock Humansplat: Generalizable single-image human gaussian splatting with structure priors.
\newblock \emph{Advances in Neural Information Processing Systems}, 37:\penalty0 74383--74410, 2024.

\bibitem[Qian et~al.(2024)Qian, Wang, Mihajlovic, Geiger, and Tang]{3dgs}
Zhiyin Qian, Shaofei Wang, Marko Mihajlovic, Andreas Geiger, and Siyu Tang.
\newblock 3dgs-avatar: Animatable avatars via deformable 3d gaussian splatting.
\newblock In \emph{Proceedings of the IEEE/CVF conference on computer vision and pattern recognition}, pages 5020--5030, 2024.

\bibitem[Song et~al.(2021)Song, Wei, Li, Liu, and Lin]{3dpt}
Chaoyue Song, Jiacheng Wei, Ruibo Li, Fayao Liu, and Guosheng Lin.
\newblock 3d pose transfer with correspondence learning and mesh refinement.
\newblock \emph{Advances in Neural Information Processing Systems}, 34:\penalty0 3108--3120, 2021.

\bibitem[Su et~al.(2021)Su, Yu, Zollh{\"o}fer, and Rhodin]{anerf}
Shih-Yang Su, Frank Yu, Michael Zollh{\"o}fer, and Helge Rhodin.
\newblock A-nerf: Articulated neural radiance fields for learning human shape, appearance, and pose.
\newblock \emph{Advances in neural information processing systems}, 34:\penalty0 12278--12291, 2021.

\bibitem[{Truebones Motions Animation Studios}()]{zoo}
{Truebones Motions Animation Studios}.
\newblock {The Amazing Animals Zoo Dataset}.

\bibitem[Weng et~al.(2022)Weng, Curless, Srinivasan, Barron, and Kemelmacher-Shlizerman]{humannerf}
Chung-Yi Weng, Brian Curless, Pratul~P Srinivasan, Jonathan~T Barron, and Ira Kemelmacher-Shlizerman.
\newblock Humannerf: Free-viewpoint rendering of moving people from monocular video.
\newblock In \emph{Proceedings of the IEEE/CVF conference on computer vision and pattern Recognition}, pages 16210--16220, 2022.

\bibitem[Xie et~al.(2024)Xie, Zong, Qiu, Li, Feng, Yang, and Jiang]{physgaussian}
Tianyi Xie, Zeshun Zong, Yuxing Qiu, Xuan Li, Yutao Feng, Yin Yang, and Chenfanfu Jiang.
\newblock Physgaussian: Physics-integrated 3d gaussians for generative dynamics.
\newblock In \emph{Proceedings of the IEEE/CVF Conference on Computer Vision and Pattern Recognition}, pages 4389--4398, 2024.

\bibitem[Xiu et~al.(2023)Xiu, Yang, Cao, Tzionas, and Black]{econ}
Yuliang Xiu, Jinlong Yang, Xu Cao, Dimitrios Tzionas, and Michael~J Black.
\newblock Econ: Explicit clothed humans optimized via normal integration.
\newblock In \emph{Proceedings of the IEEE/CVF conference on computer vision and pattern recognition}, pages 512--523, 2023.

\bibitem[Xu et~al.(2020{\natexlab{a}})Xu, Bazavan, Zanfir, Freeman, Sukthankar, and Sminchisescu]{ghum}
Hongyi Xu, Eduard~Gabriel Bazavan, Andrei Zanfir, William~T Freeman, Rahul Sukthankar, and Cristian Sminchisescu.
\newblock Ghum \& ghuml: Generative 3d human shape and articulated pose models.
\newblock In \emph{Proceedings of the IEEE/CVF Conference on Computer Vision and Pattern Recognition}, pages 6184--6193, 2020{\natexlab{a}}.

\bibitem[Xu et~al.(2021)Xu, Alldieck, and Sminchisescu]{hnerf}
Hongyi Xu, Thiemo Alldieck, and Cristian Sminchisescu.
\newblock H-nerf: Neural radiance fields for rendering and temporal reconstruction of humans in motion.
\newblock \emph{Advances in Neural Information Processing Systems}, 34:\penalty0 14955--14966, 2021.

\bibitem[Xu et~al.(2020{\natexlab{b}})Xu, Zhou, Kalogerakis, Landreth, and Singh]{rignet}
Zhan Xu, Yang Zhou, Evangelos Kalogerakis, Chris Landreth, and Karan Singh.
\newblock Rignet: Neural rigging for articulated characters.
\newblock \emph{arXiv preprint arXiv:2005.00559}, 2020{\natexlab{b}}.

\bibitem[Yang et~al.(2021{\natexlab{a}})Yang, Sun, Jampani, Vlasic, Cole, Chang, Ramanan, Freeman, and Liu]{lasr}
Gengshan Yang, Deqing Sun, Varun Jampani, Daniel Vlasic, Forrester Cole, Huiwen Chang, Deva Ramanan, William~T Freeman, and Ce Liu.
\newblock Lasr: Learning articulated shape reconstruction from a monocular video.
\newblock In \emph{CVPR}, 2021{\natexlab{a}}.

\bibitem[Yang et~al.(2021{\natexlab{b}})Yang, Sun, Jampani, Vlasic, Cole, Liu, and Ramanan]{viser}
Gengshan Yang, Deqing Sun, Varun Jampani, Daniel Vlasic, Forrester Cole, Ce Liu, and Deva Ramanan.
\newblock Viser: Video-specific surface embeddings for articulated 3d shape reconstruction.
\newblock In \emph{NeurIPS}, 2021{\natexlab{b}}.

\bibitem[Yang et~al.(2022)Yang, Vo, Neverova, Ramanan, Vedaldi, and Joo]{banmo}
Gengshan Yang, Minh Vo, Natalia Neverova, Deva Ramanan, Andrea Vedaldi, and Hanbyul Joo.
\newblock Banmo: Building animatable 3d neural models from many casual videos.
\newblock In \emph{CVPR}, 2022.

\bibitem[Zakharkin et~al.(2021)Zakharkin, Mazur, Grigorev, and Lempitsky]{point}
Ilya Zakharkin, Kirill Mazur, Artur Grigorev, and Victor Lempitsky.
\newblock Point-based modeling of human clothing.
\newblock In \emph{Proceedings of the IEEE/CVF International Conference on Computer Vision}, pages 14718--14727, 2021.

\bibitem[Zhang et~al.(2024{\natexlab{a}})Zhang, Chang, Li, Soleymani, and Ahuja]{mp4d}
Hao Zhang, Di Chang, Fang Li, Mohammad Soleymani, and Narendra Ahuja.
\newblock Magicpose4d: Crafting articulated models with appearance and motion control.
\newblock \emph{arXiv preprint arXiv:2405.14017}, 2024{\natexlab{a}}.

\bibitem[Zhang et~al.(2024{\natexlab{b}})Zhang, Li, and Ahuja]{opennerf}
Hao Zhang, Fang Li, and Narendra Ahuja.
\newblock Open-nerf: Towards open vocabulary nerf decomposition.
\newblock In \emph{Proceedings of the IEEE/CVF Winter Conference on Applications of Computer Vision}, pages 3456--3465, 2024{\natexlab{b}}.

\bibitem[Zhang et~al.(2024{\natexlab{c}})Zhang, Li, Rawlekar, and Ahuja]{learning}
Hao Zhang, Fang Li, Samyak Rawlekar, and Narendra Ahuja.
\newblock Learning implicit representation for reconstructing articulated objects.
\newblock \emph{arXiv preprint arXiv:2401.08809}, 2024{\natexlab{c}}.

\bibitem[Zhang et~al.(2024{\natexlab{d}})Zhang, Li, Rawlekar, and Ahuja]{s3o}
Hao Zhang, Fang Li, Samyak Rawlekar, and Narendra Ahuja.
\newblock S3o: A dual-phase approach for reconstructing dynamic shape and skeleton of articulated objects from single monocular video.
\newblock In \emph{International Conference on Machine Learning}, pages 59191--59209. PMLR, 2024{\natexlab{d}}.

\bibitem[Zhang et~al.(2024{\natexlab{e}})Zhang, Yu, Wu, Feng, Zheng, Snavely, Wu, and Freeman]{physdreamer}
Tianyuan Zhang, Hong-Xing Yu, Rundi Wu, Brandon~Y Feng, Changxi Zheng, Noah Snavely, Jiajun Wu, and William~T Freeman.
\newblock Physdreamer: Physics-based interaction with 3d objects via video generation.
\newblock In \emph{European Conference on Computer Vision}, pages 388--406. Springer, 2024{\natexlab{e}}.

\bibitem[Zhao et~al.(2024)Zhao, Wang, Zhao, Wang, Wu, Long, and Zou]{simany}
Haoyu Zhao, Hao Wang, Xingyue Zhao, Hongqiu Wang, Zhiyu Wu, Chengjiang Long, and Hua Zou.
\newblock Automated 3d physical simulation of open-world scene with gaussian splatting, 2024.

\end{thebibliography}
}

\clearpage \appendix \section{Appendix Section}
\label{appendix}
\begin{figure*}[ht]
    \centering
    \includegraphics[width=1\textwidth]{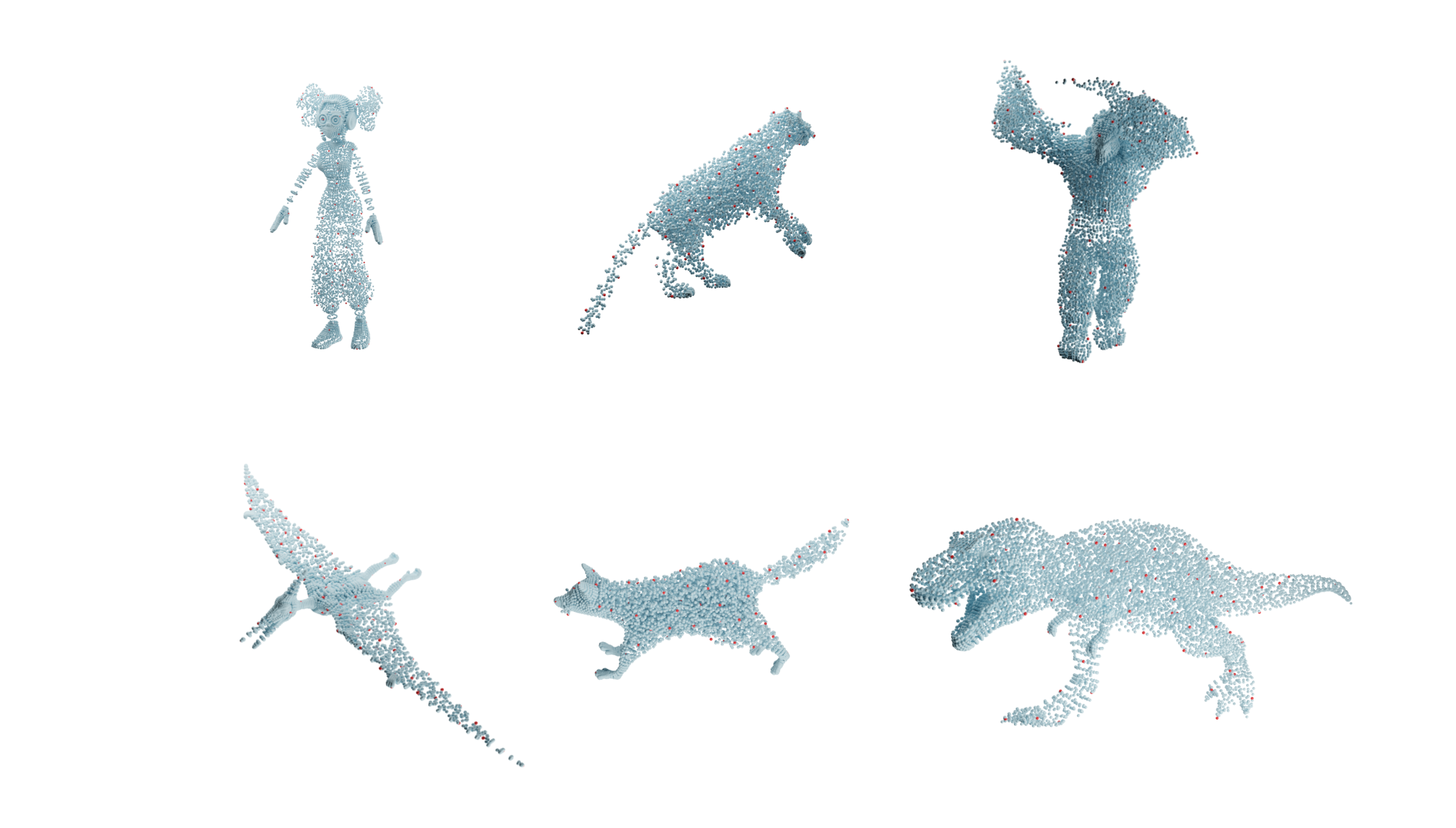}
    \caption{Visualization of Material Prototype Centers.}
    \label{mpc}
\end{figure*}

Due to page limitations in the main text, we have included additional supplementary materials in the appendix. This section primarily provides the following:

\begin{enumerate}
    \item A comprehensive introduction to our newly proposed dataset (\textbf{Sec. A.1}).
    \item A detailed description of the \textbf{implementation} (\textbf{Sec. A.2}).
    \item A qualitative comparison and analysis of \textbf{inverse skinning} using neural blend skinning weights (\textbf{Sec. A.3}).
    \item Setup for User Study (\textbf{Sec. A.4}).
    \item Visualization of Material Prototype Centers (\textbf{Fig.\ref{mpc}}).
\end{enumerate}

\subsection{Dataset}  
\label{dataset}
To evaluate PhysRig and compare its performance against traditional Linear Blend Skinning (LBS) methods, we construct a diverse simulation dataset tailored for the inverse skinning task. This dataset enables a comprehensive analysis of PhysRig's ability to recover underlying motion parameters and material properties under various challenging conditions.  

We curate 17 structurally distinct objects from Objaverse, Mixamo, and The Amazing Animals Zoo, ensuring broad coverage of different articulation types and deformation patterns. These objects are categorized into three groups:  
(i) Humanoid Characters (5 objects), (ii) Quadruped Animals (6 objects): leopard, mammoth, stego, krin, cow, and raccoon, and (iii) Other Entities (6 objects): t-rex, pterosaur, whale, angelfish, cobra, and shark    
Each object is associated with 1 to 4 motion sequences, resulting in a total of 40 motion sequences, with each sequence containing 20 to 100 frames. This setup ensures a diverse range of temporal dynamics, allowing us to evaluate PhysRig's generalization across various topologies and articulation mechanisms.  

To further assess PhysRig's ability to learn material properties, we provide two different material configurations for each of the 40 motion sequences, resulting in a total of 80 cases:  
(i) Homogeneous-material objects, where the entire structure exhibits a uniform material property.  
(ii) Heterogeneous-material objects, where different regions are assigned distinct material properties, simulating realistic soft-tissue variations and composite structures.  
In total, our dataset consists of 120 cases, including 40 cases with original motion sequences and 80 cases with different material configurations. By systematically introducing controlled material variations, our dataset enables a fine-grained evaluation of PhysRig's capability to recover Young’s modulus, Poisson’s ratio, and skeletal motion parameters across diverse material configurations. Given that the objects in our dataset primarily consist of animals and humans, which typically exhibit similar Poisson’s ratios across different tissues, we assume a homogeneous Poisson’s ratio for all objects.  
This dataset serves as a quantitative benchmark, evaluating both motion accuracy and material property estimation.

\begin{figure*}[ht]
    \centering
    \includegraphics[width=1\textwidth]{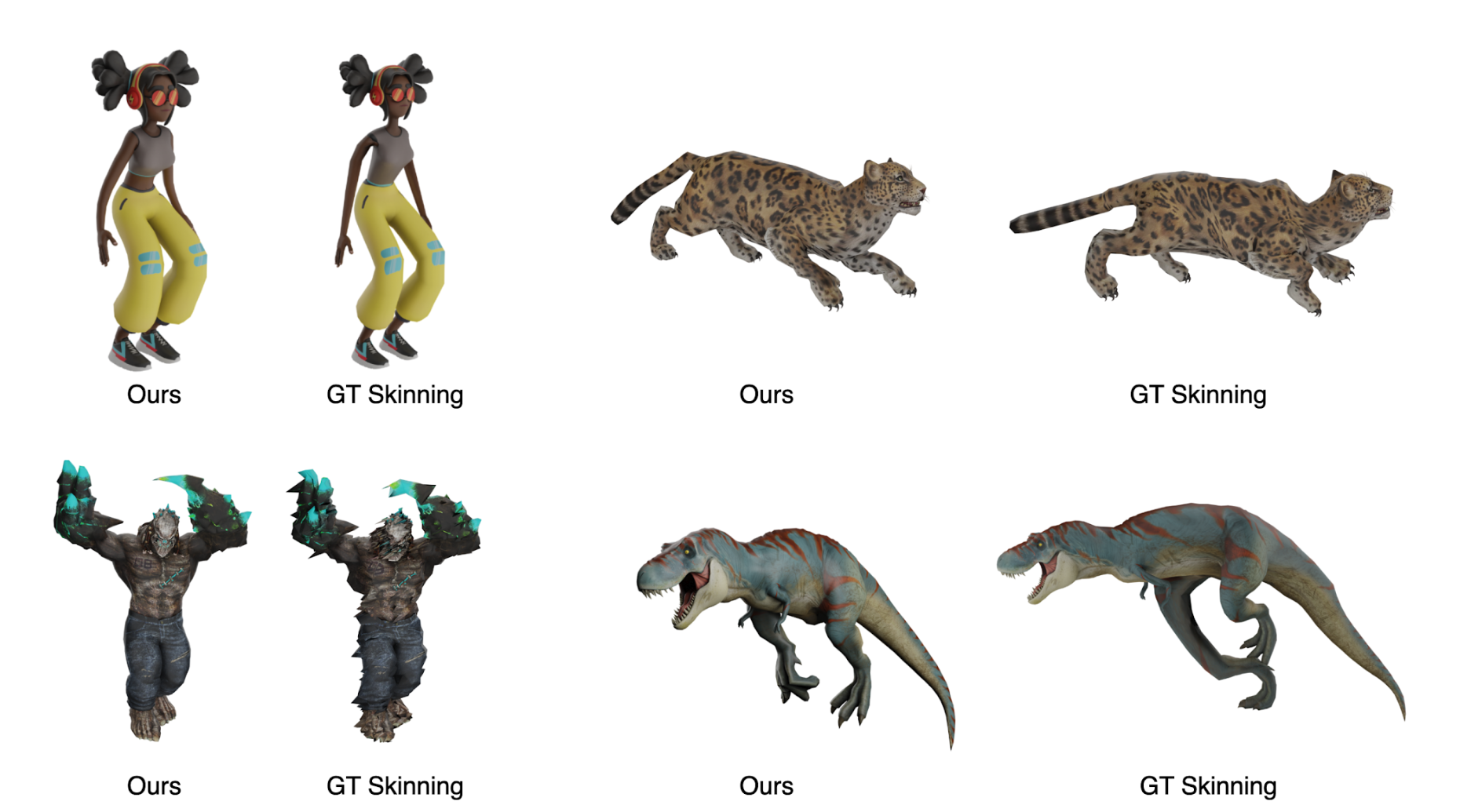}
    \caption{Qualitative Comparisons of \textbf{PhysRig} and neural linear blend skinning using ground truth skinning weights as initialization.}
    \label{exp_3}
\end{figure*}

\subsection{Implementation Details}

In our experiments, the mesh object consists of approximately 2000 to 50000 vertices. We discretize the simulation field into a \(100^3\) grid for simulation. To reduce computational complexity, the number of material prototypes is set to be 25–200, which are uniformly distributed within the volume. For accurate motion modeling, we employ 100 sub-steps between successive frames (25 FPS), corresponding to a duration of:
\[
\Delta t = 4 \times 10^{-4} \text{ seconds per sub-step}.
\]
We adopt an alternating optimization strategy to separately optimize the material properties and velocity. Specifically, the material parameters are optimized using the AdamW optimizer, while the velocity parameters are optimized using SGD. To achieve efficient convergence, the initial learning rate for material training is set to be 20 times that of velocity training. The initial learning rate for velocity optimization is adjusted based on different scenarios, ranging from \(5 \times 10^{-3}\) to \(2 \times 10^{-2}\), with commonly used values of 0.008 and 0.01. The material learning rate is set accordingly as 20 times the velocity learning rate. All optimization processes employ linear learning rate decay, with the learning rate reset to its initial value at the beginning of each alternating phase. During training, each scene undergoes three alternating optimization cycles for material and velocity, where the material is trained for 20 iterations per cycle, while the velocity is optimized for 30 iterations per frame. This iterative alternating optimization strategy gradually reduces the coupling between velocity and material, leading to more stable parameter learning.

\textbf{Voxel-Based Adaptive Sampling.}
To achieve a comprehensive volumetric representation, we first load the input triangular mesh and extract its vertices and surface sample points. We then adaptively partition the space into multiple voxels based on the bounding box of the mesh and a pre-defined resolution.
For each voxel, we evaluate its center point to determine whether it lies inside the mesh. If the center is within the mesh interior, we randomly generate a set number of sampling points inside the voxel. Each of these points is then checked to verify whether it remains inside the mesh. Finally, we aggregate all valid interior sampling points, along with surface points and original vertices, into a unified complete point cloud. This ensures a dense and well-distributed volumetric representation of the input mesh, which is subsequently outputted for further processing.

\subsection{Qualitative Comparisons on Inverse Skinning}
Although the LBS-based method is initialized with ground truth skinning weights, we observe that when jointly optimizing skinning weights and bone transformations, the optimization process can fall into a suboptimal local minimum due to incorrect bone transformations, ultimately leading to unsatisfactory results. Additionally, despite incorporating the least motion loss, LBS often exhibits noticeable frame-to-frame jitter or abrupt changes in motion.  

Moreover, LBS can suffer from various artifacts, such as unrealistic folding (e.g., in the leopard case), incorrect changes in volume size (e.g., in the Michelle case), or distorting the wrong body parts in an attempt to minimize loss, leading to severe deformations (e.g., in the T-Rex case).  

In contrast, PhysRig, being grounded in a physics-based simulator, produces significantly more realistic and physically plausible results.

\subsection{User Study}
\label{user study}
 We provide a user study for comparison between PhysRig and neural linear blend skinning on inverse skinning. We asked 50 lay participants from online platforms for user studies, to rate the quality of \textbf{120} optimized mesh sequences from LBS-1, 2, 3, and PhysRig on a scale of 0 (low) to 5 (high). The participants are paid at an hourly rate of 16 USD. The results from different methods are anonymized as A to E, and the order is randomized. We provide the video visualization of these comparisons.

\ifarxiv \clearpage \appendix  \fi

\end{document}